# A Comparative Evaluation of Large Language Models for Persian Sentiment Analysis and Emotion Detection in Social Media Texts

Kian Tohidi[1], Kia Dashtipour[2], Simone Rebora[1], Sevda Pourfaramarz[1]

[1] University of Verona
[2] Edinburgh Napier University

**Abstract**

This study presents a comprehensive comparative evaluation of four state-of-the-art Large Language Models (LLMs)—Claude 3.7 Sonnet, DeepSeek-V3, Gemini 2.0 Flash, and GPT-4o—for sentiment analysis and emotion detection in Persian social media texts. Comparative analysis among LLMs has witnessed a significant rise in recent years, however, most of these analyses have been conducted on English language tasks, creating gaps in understanding cross-linguistic performance patterns. This research addresses these gaps through rigorous experimental design using balanced Persian datasets containing 900 texts for sentiment analysis (positive, negative, neutral) and 1,800 texts for emotion detection (anger, fear, happiness, hate, sadness, surprise). The main focus was to allow for a direct and fair comparison among different models, by using consistent prompts, uniform processing parameters, and by analyzing the performance metrics such as precision, recall, F1-scores, along with misclassification patterns. The results show that all models reach an acceptable level of performance, and a statistical comparison of the best three models indicates no significant differences among them. However, GPT-4o demonstrated a marginally higher raw accuracy value for both tasks, while Gemini 2.0 Flash proved to be the most cost-efficient. The findings indicate that the emotion detection task is more challenging for all models compared to the sentiment analysis task, and the misclassification patterns can represent some challenges in Persian language texts. These findings establish performance benchmarks for Persian NLP applications and offer practical guidance for model selection based on accuracy, efficiency, and cost considerations, while revealing cultural and linguistic challenges that require consideration in multilingual AI system deployment.

## 1. Introduction

The swift progress in artificial intelligence along with natural language processing technologies has entirely reshaped our methods for interpreting human expression in digital media. Sentiment analysis and emotion detection represent key applications in this field, involving separate but related computational tasks that extract emotional information from written language. Sentiment analysis determines whether text expresses positive, negative or neutral sentiment while emotion detection explores more specific feelings including anger, fear, sadness, and surprise (Wankhade et al., 2022; Nandwani & Verma, 2021; Babu & Kanaga, 2022).

The development of Large Language Models (LLMs) transformed these fields through their advanced abilities to grasp contextual subtleties and linguistic intricacies which conventional methods found difficult to handle. LLMs showcase the leading edge of natural language understanding capabilities while achieving outstanding performance in multiple languages and domains. These models surpass older approaches by functioning well in zero-shot and few-shot learning situations which allows them to handle new tasks without needing significant task-specific training (Manning et al., 2008). Large pre-trained language models like BERT and GPT have significantly improved their sentiment analysis performance by training on large datasets, which represent rich contextual features of the elements of language (Gupta et al., 2024).

However, research efforts and evaluation of these models have mainly focused on English applications because of the ready access to large English datasets and the concentration of research resources within English-speaking regions. This situation creates gaps in understanding cross-linguistic performance patterns and limits our knowledge of how these powerful models perform in various linguistic environments that possess unique structural and cultural features. The field of sentiment analysis and emotion detection of social media has been the focus of significant interest from both academic and industrial sectors, impelled by manifold applications and the accessibility of data from such popular websites as X, Facebook, and Instagram (Rodríguez-Ibánez et al., 2023).

Despite significant advances in LLM capabilities, comprehensive evaluation of their performance across different languages remains limited (Zhang et al., 2023). The majority of comparative studies focus on English-language applications which restricts understanding of performance patterns across different linguistic settings. The limitation becomes especially significant for languages which possess distinctive structural features and cultural communication styles and therefore present unique obstacles for automated analytical tools. The current body of research demonstrates multiple significant limitations that need addressing. Most comparative research into advanced language models concentrates on English language scenarios which restricts understanding of performance differences among various linguistic settings. Most research efforts analyze individual models separately or perform limited comparisons which blocks researchers from understanding the full range of strengths and practical compromises between various methods. Standardized evaluation frameworks that assess cross-linguistic sentiment analysis and emotion detection have not yet reached full development, making it difficult for researchers to reach meaningful conclusions regarding model capabilities when working with diverse languages and cultural backgrounds.

Persian language processing presents particular challenges due to its unique structural and cultural characteristics. Persian serves as the official language across Iran, Tajikistan, and Afghanistan, with over one hundred million speakers globally (Basiri et al., 2014), yet requires specialized natural language processing techniques because its unique features create substantial challenges for computational linguistics compared to Western languages. The Persian writing system uses a 32-letter alphabet that flows from right to left, displays considerable complexity because each letter shows three to four different shapes according to its position in words, and presents distinctive grammatical structures and syntactic configurations which produce irregular word ordering patterns (Shamsfard, 2011; Rajabi & Valavi, 2021). Persian language morphology exhibits complex variations through its use of multiple verbal suffix types and extensive declensional suffix changes together with several methods of forming plurals (Rajabi & Valavi, 2021). The language includes Arabic-derived phonetic elements which writers can choose to represent in text thus allowing different valid orthographic forms for single words. Persian utilizes half-space characters for specific grammatical constructions, yet users consistently omit these markers in opinion texts, creating additional tokenization challenges that directly impact sentiment analysis accuracy (Rajabi & Valavi, 2021). Sentiment analysis in Persian encounters specific cultural difficulties because the same words demonstrate contrasting polarity across various Persian-speaking communities where they can signify positive meaning in one context but translate to negative or neutral in another (Dehkharghani, 2019).

This study aims to fill this gap by conducting a systematic comparative analysis of LLMs in Persian language contexts and is focused on addressing the following four key research questions: What are the accuracy levels of Claude 3.7 Sonnet, DeepSeek-V3, Gemini 2.0 Flash, and GPT-4o in classifying the sentiment polarity (positive, negative, neutral) of Persian social media texts? How do the four selected LLMs compare in their ability to detect specific emotions (anger, fear, happiness, hate, sadness, surprise) in Persian language social media texts? What are the trade-offs between accuracy, computational efficiency, and cost-effectiveness among these models for Persian text analysis tasks? What are the most common misclassification patterns exhibited by each model, and what do these patterns reveal about the challenges of Persian sentiment and emotion analysis?

The objective of this study is a systematic comparative evaluation based on standard datasets, established evaluation protocols, and performance metrics. The study adheres to principles of controlled experimentation by employing balanced dataset sampling and consistent prompt engineering complemented by a comprehensive error analysis. The present work has several contributions to computational linguistics and the field of Persian language technology. A key contribution is the comprehensive comparative evaluation of state-of-the-art LLMs for Persian sentiment analysis and emotion detection tasks, which addresses a gap in cross-linguistic evaluation research. The research establishes a robust evaluation framework that can be tailored for other languages and diverse cultural contexts through the standardization of dataset preparation, prompt engineering, and performance analysis. This study provides practical recommendations for developers of Persian-language applications, by considering accuracy metrics, computational efficiency, and cost. The inclusion of this practical approach helps to bridge the gap between research findings and real-world deployment needs. The detailed analysis of misclassification patterns contributes to theoretical understanding of how modern language models handle linguistic and cultural variation, providing insights that extend beyond the specific models evaluated and establishing performance benchmarks for Persian NLP applications while offering practical guidance for model selection based on accuracy, efficiency, and cost considerations.

## 2. Related Work

The use of Large Language Models (LLMs) in sentiment analysis and emotion detection is a fast growing research field. However, there are a limited number of systematic comparative analyses of this research area for different languages. In this section, the literature on LLM-based sentiment analysis, recent advancements in social media text analysis, and previous research on emotion detection in the Persian language are reviewed to contextualize the current research.

## 2.1 Large Language Models in Sentiment Analysis

Studies have shown that LLMs have demonstrated their effectiveness in performing different natural language processing tasks with considerable success. A common notable observation is that LLMs exhibit zero-shot and few-shot in-context learning ability that provides an adequate level of performance even without prior supervised training (Bang et al., 2023; Ye et al., 2023; Zhong et al., 2023; Yang et al., 2024). On the other hand, an extensive evaluation of LLM performance on the sentiment analysis task suggests task-dependent variations, which needs to be taken into account in practical settings.

Zhang et al. (2023) performed a systematic study of LLMs on sentiment analysis tasks and found that performance depends on the complexity of the task. For simpler sentiment analysis tasks like binary or trinary classification tasks, LLMs generally exhibit strong zero-shot capabilities without any need for prior training. The study showed that ChatGPT performed comparably to the T5 model, in particular fine-tuned on complete training sets for each individual dataset. On simple classification tasks, ChatGPT's performance reached 97% of T5's prediction accuracy. The paper also reports a score of 83% of T5's prediction accuracy for Multifaceted Analysis of Subjective Text (MAST) tasks. However, in more complex tasks that required a deeper level of analysis, LLMs had a weaker performance relative to fine-tuned models, with scores of 52.4 versus 65.6 accuracy on Yelp-5 datasets for fine-grained five-class sentiment classification and 72.80 versus 80.35 on tasks involving comparative opinion mining.

The performance gap is more significant in the emotion detection task because the detection task has a greater number of categories and smaller differences between emotions, which makes it more difficult for zero-shot studies. This result directly relates to the current study's examination of Persian emotion detection in six different emotion categories, suggesting that LLMs' performance varies in sentiment analysis and emotion detection tasks.

Along with their remarkable capabilities, LLMs encounter several critical issues like bias, privacy, and transparency concerns that influence their application in practical tasks (Ranjan et al., 2024; Liyanage & Ranaweera, 2023). These models can memorize and reinforce biases present in the training data, leading to gender, racial, and contextual biases which can potentially result in discrimination (Bolukbasi et al., 2016; Caliskan et al., 2017; Binns, 2018).

## 2.2 Recent Advances in Social Media Sentiment Analysis

Contemporary research in social media sentiment analysis has focused on developing more sophisticated approaches that leverage LLM capabilities while addressing domain-specific challenges. Krugmann and Hartmann (2024) performed an extensive evaluation of LLMs for sentiment analysis tasks, comparing their performance to state-of-the-art transfer learning models across three key experiments. In zero-shot settings, the performance of LLMs on binary and three-class sentiment classification, across more than 3,900 texts drawn from 20 datasets, was found to match or outperform prior transfer learning approaches. The research also found that certain features of the input text such as dataset source, the number of sentences and words, and linguistic features had a noticeable impact on performance. These include dataset provenance and the presence of a neutral class as factors for reduced accuracy, and longer texts and content-dominant words as improving it.

He et al. (2024) extended this study to compare the performance of different prompting and fine-tuning techniques for sentiment analysis of social media data on healthcare topics. In particular, their analyses of posts related to Healthcare Reform, vaccination, mask-wearing, and quality of healthcare services showed that the performance of LLMs was better than that of VADER, yet these LLMs were still unable to accurately assign correct sentiment labels in many cases. Results indicated that performance increased when using data-specific prompts that incorporated both contextual and task-related target information, highlighting the potential utility of prompt engineering approaches for domain-specific tasks.

Addressing the increasing relevance of cross-lingual sentiment analysis, Miah et al. (2024) proposed an ensemble system that integrates transformer models with LLMs to enhance sentiment analysis performance by translation-based methods. The experimental results showed more than 86% accuracy for translated sentences and outperformed the stand-alone pre-trained models and LLMs. However, translation-based methods struggle with accuracy and processing cultural contexts, idiomatic expressions, and colloquial terms.

The recent study also focused on the multimodal aspect of sentiment analysis. Al-Tameemi et al. (2024) reported that the importance of multimodal sentiment analysis is also emerging in social media. The researchers pointed out that the traditional sentiment analysis is limited to textual data which could not explain all aspects of the communication among people of the different media types. The multimodal sentiment analysis involves using both visual and textual information to gather more accurate understanding of users' opinions by using the various sources of information which is often more effective than single-mode methods.

## 2.3 Persian Language Sentiment Analysis and Emotion Detection

Persian sentiment analysis and emotion detection research has encountered unique challenges due to Persian language structural features, scarcity of annotated datasets, and cultural complexities in emotional expression. Recent efforts have focused on developing high-quality annotated datasets and advancing methodological approaches specifically tailored for Persian text analysis.

Mirzaee et al. (2025) made substantial contributions to Persian emotion detection research through creating ArmanEmo, containing over 7,000 labeled Persian sentences from various social media sources including X, Instagram, and Digikala comments. The dataset utilizes Ekman's six primary emotions—anger, fear, happiness, hatred, sadness, and surprise—while adding a seventh "Other" category to encompass expressions outside basic emotions. Their thorough annotation procedure selected 12 annotators from 35 candidates, guaranteeing high-quality labeling through consensus requirements and validated test set assessments.

Complementing this broader emotional scope, Rasouli and Kiani (2023) developed ShortPersianEmo, a specialized dataset focusing exclusively on short Persian texts containing 5,472 texts ranging from 30 to 100 words collected from X and Digikala platforms. Using Rachael Jack's emotional model, texts were categorized into five emotional classes: happiness, sadness, anger, fear, and other. This attention has a strong practical motivation as many social media texts are short, making the emotion detection task more challenging for such inputs.

Persian emotion detection research has evolved from traditional lexicon-based approaches toward sophisticated deep learning methodologies, with transfer learning emerging as particularly effective. Initial studies focused on lexicon-based methods using resources such as the NRC Word-Emotion Association Lexicon, however, such methods did not achieve impressive results because they were incapable of handling contextualized information, and culture-specific emotion words (Mirzaee et al., 2025). One important shift was the move from shallow learning to deep learning methods. The work of Rasouli and Kiani (2023) highlights the significant performance gap between these shallow learning and deep learning approaches when using both context-free and context-aware embeddings.

The development of Persian-specific language models has shown to significantly improve emotion detection performance. ParsBERT, a monolingual Persian BERT-based language model, has been outperforming multilingual models in a number of studies. Mirzaee et al. (2025) showed that XLM-RoBERTa-large achieved the highest accuracy (macro-averaged F1 score of 75.39%) across a number of models while Rasouli and Kiani (2023) reported Deep ParsBERT outperformed context-free FastText embeddings by 9% (73% overall accuracy) on short Persian text samples while also outperforming shallow learning approaches by a large margin.

A series of cross-dataset studies has also been conducted to investigate Persian emotion detection systems' generalizability and robustness. Mirzaee et al. (2025) conducted transfer learning experiments, demonstrating significant quality gap between the newly introduced ArmanEmo and existing EmoPars datasets by ArmanEmo-trained models' F1 scores outperforming the EmoPars-trained models on EmoPars test set by over 19%. The high difference in model performance was attributed to the more rigorous annotation approach in ArmanEmo collection, highlighting the importance of dataset quality in Persian emotion detection research.

Although substantial progress made, there are some essential limitations in the field of Persian emotion detection, which limit the system's performance and its generalization. The size of existing datasets is still limited in comparison to English resources, which restricts further development of sophisticated models and comprehensive testing across diverse topics and scenarios. Consistency and quality of annotations are another issue, with human annotators not being able to agree upon the label for a large part of the texts in both studies, reflecting the subjective nature of emotions and the need for more refined annotation mechanisms.

The current research environment demonstrates that cutting-edge language models receive limited evaluation through uniform assessment frameworks. The absence of standardized evaluation methods prevents researchers from making clear judgments about language model performance and optimal application strategies across various contexts. This gap in systematic comparative evaluations for state-of-the-art LLMs on Persian sentiment analysis and emotion detection in social media contexts directly motivates the comprehensive comparative study presented in this research, which addresses these limitations through rigorous experimental design and standardized evaluation protocols.

## 3. Methodology

This chapter evaluates the performance of four Large Language Models (LLMs) named Claude 3.7 Sonnet, DeepSeek-V3, Gemini 2.0 Flash, and GPT-4o through a methodological framework that uses two separate Persian datasets with one dataset dedicated to sentiment analysis and the other to emotion detection. The methodology starts with dataset selection and cleaning which then progresses to sampling and prompt design before model inference and evaluation through classification metrics. Classification metrics such as precision, recall, and F1-score are used to measure each

LLM's performance by comparing model outputs with original labeled data. All code implementations used for data processing, cleaning, and analysis in this study are publicly available in a dedicated GitHub[1] repository, to maintain transparency and enable reproducibility of the described methodology.

### 3.1 Dataset Selection and Evaluation

#### 3.1.1 Initial Dataset Collection

The research began with an initial analysis of five Persian-language datasets which were publicly accessible. The datasets differed across task focus which included sentiment analysis and emotion detection and varied in their data source which consisted of X (formerly Twitter), Instagram, and Digikala[2] as well as label quality and documentation. The first dataset, *Persian sentiment analysis dataset*[3], was available on Kaggle and contained labeled texts from Instagram. The second dataset, *Persian Twitter Dataset – Sentiment Analysis*[4], was also available on Kaggle and included labeled texts from X. The third dataset, *ArmanEmo*[5], was available on GitHub and consisted of labeled texts from X, Instagram, and Digikala. The fourth dataset, *Persian-Emotion-Detection*[6], was also available on GitHub and contained labeled texts from X. Finally, the fifth dataset, *ShortPersianEmo*[7], was also available on GitHub and included labeled texts from X and Digikala.

#### 3.1.2 Sampling and Manual Accuracy Evaluation

A manual inspection was conducted to determine the most suitable datasets for this research (see Table 1). A random selection of 150 texts was made from each dataset and a native Persian speaker checked the labeled data for accuracy and consistency. The evaluation provided important details about dataset quality yet is subjective and limited by the small sample size. Various annotators might have applied different interpretations to particular labels and increasing the sample size might expose more inconsistencies.

|  | Size | Type | Labeling Accuracy Count | Labeling Accuracy Percentage | Published Research Paper | Number of Annotators | Requires Cleaning |
|---|---|---|---|---|---|---|---|
| Dataset 1 | 8,512 | Sentiment Analysis | 136 out of 150 | 90.66 % | No | Unknown | Yes |
| Dataset 2 | 3,381 | Emotion Detection | 131 out of 150 | 87.33 % | Yes | 2 | Yes |
| Dataset 3 | 7,276 | Emotion Detection | 145 out of 150 | 96.66 % | Yes | 12 | Yes |
| Dataset 4 | 30,000 | Emotion Detection | 92 out of 150 | 61.33 % | Yes | 5 | Yes |
| Dataset 5 | 5,472 | Emotion Detection | 141 out of 150 | 94 % | Yes | 5 | Yes |

*Table 1 – Evaluation of the 5 Persian Datasets*

For this research, having a high-accuracy dataset was prioritized above all other factors like documentation or annotation transparency since experimental validity relies directly on input label reliability. Among the five datasets evaluated, three were excluded due to significant limitations that would compromise experimental validity. Dataset 2 demonstrated major class imbalance with only 3,381 labeled texts, containing 1,120 texts marked as "sad" versus 388 marked as "angry," and included 415 texts categorized as "intense emotions" which were not clearly interpretable. Dataset 4, despite being the largest with 30,000 labeled texts, demonstrated the worst labeling consistency with approximately 80% of texts lacking emotion labels, and the remaining entries carried inconsistent multiple labels that created difficulties in interpretation. Dataset 5, while matching Dataset 3 in accuracy, contained fewer samples with 5,472 texts and exhibited significant emotion imbalance with 1,625 samples labeled as "happy" compared to only 380 labeled "fear," additionally requiring more extensive cleaning procedures due to prevalent Digikala entries that increased the risk of processing errors.

---

[1] https://github.com/KianTohidi/Persian_Sentiment_and_Emotion
[2] An Iranian e-commerce company.
[3] https://www.kaggle.com/datasets/instatext/persian-sentiment-analysis-dataset (accessed April 2025).
[4] https://www.kaggle.com/datasets/mohammadalimkh/persian-twitter-dataset-sentiment-analysis (accessed April 2025). Related paper: https://doi.org/10.48550/arXiv.2504.10662
[5] https://github.com/Arman-Rayan-Sharif/arman-text-emotion (accessed April 2025). Related paper: https://doi.org/10.1007/s10579-025-09817-4.
[6] https://github.com/nazaninsbr/Persian-Emotion-Detection/tree/main (accessed April 2025). Related paper: http://dx.doi.org/10.26615/issn.2603-2821.2021_023.
[7] https://github.com/vkiani/ShortPersianEmo/tree/main (accessed April 2025). Related paper: https://doi.org/10.22044/jadm.2023.13675.2485.

Based on the assessment described above, Dataset 1 and Dataset 3 were selected for further analysis, and they will be called from this moment on as the Sentiment dataset and the Emotion dataset, respectively. The Sentiment dataset was selected for its high labeling accuracy, balanced sentiment classes, and minimal cleaning requirements, and proved to be ideal for sentiment analysis. Although the Emotion dataset had few limitations it was chosen for emotion analysis because of its excellent labeling quality along with a substantial number of labeled entries and a balanced distribution of emotion classes.

### 3.2 Dataset Cleaning and Preprocessing

#### 3.2.1 Cleaning Process of the Sentiment Dataset

The Sentiment dataset underwent several preprocessing and cleaning steps in the Google Colab via the Pandas library to set it up for sentiment analysis. To achieve accurate Persian text representation, the dataset was loaded with "utf-8-sig" encoding. The sentiment labels were also converted from numerical values to categorical ones: The label "1" became "positive", "0" turned into "neutral", and "-1" transformed into "negative". The dataset label conversion was essential for matching with LLMs expected outputs so that evaluation results could be directly compared.
Duplicate records were eliminated during the cleaning stage which decreased the total number of records from 8,512 to 8,502. A minimum text length of 10 characters was applied to filter out overly short or uninformative entries, to avoid noise in sentiment classification. The dataset size decreased to 7,826 samples after this additional filtering process. Finally, the dataset was further refined when all entries exceeding 100 words were removed to eliminate inconsistent interpretations which led to 147 more entries being deleted. The final dataset size was 7,679 entries after applying this last cleaning. In total, 833 entries were removed. After cleaning, the final distribution was 39.6% "negative," 27% "neutral," and 33.4% "positive."

#### 3.2.2 Cleaning Process of Emotion Dataset

Google Colab served as the platform where the Pandas library enabled the implementation of multiple preprocessing and cleaning steps to prepare the Emotion dataset for emotion detection. Two TSV (Tab-Separated Values) files for input were loaded and they were converted to CSV files by "utf-8-sig" encoding for Persian text and displaying problems and avoiding decoding errors.
After verifying the number of rows in the two datasets, the files were merged to a one new file for unified processing and faster data cleaning. The new merged dataset contained all initial records consist of 7,276 rows. All entries have text without missing or empty text values, therefore none of the entries were removed. The dataset underwent deduplication to maintain unique entries while preventing any specific entry from being overrepresented which resulted in a reduction of 48 rows. To improve data quality, the filtering process removed 104 rows of uninformative or incomplete entries by excluding entries less than 10 characters long. Similarly, entries over 100 words were discarded because they introduced interpretation inconsistencies which resulted in two more entries being removed.
Next, the remaining emotion labels were standardized to achieve uniform terminology. The dataset labels were transformed to lowercase, and any additional spaces were stripped to ensure consistency. Adjective-like labels such as "sad," "happy," and "angry" were mapped to their corresponding noun forms: "sadness," "happiness," and "anger." Following the standardization process the dataset included seven emotion categories which were "anger," "fear," "happiness," "hate," "other," "sadness," and "surprise."
I preserved the label "hate" throughout my analyses to honor the original dataset and its labeling scheme, while acknowledging this deviation from Ekman's original classification. Ekman (1992) identified anger, disgust, fear, happiness, sadness, and surprise as the six fundamental emotions while the ArmanEmo dataset created by Mirzaee et al. (2025) replaces "disgust" with "hate." The authors make this substitution in their labels based on Ekman's model but without providing any explanation for the change.
The dataset was enhanced through the elimination of 1,820 entries marked as "other," an ambiguous category, to ensure all remaining data had precise emotion labels. After completing this cleaning process, the dataset was reduced to 5,302 unique, high-quality entries.
Two LLMs, "Claude 3.7 Sonnet," and "GPT-4o" were used to flag potential e-commerce content. They applied a shared set of e-commerce discourse markers, including references to Digikala, technical or product descriptions, and review-style language involving price, delivery, or customer service. A native Persian speaker acted as the final reviewer, which included resolving models' disagreements and ensuring proper consideration of cultural and contextual nuances.

After excluding 275 e-commerce reviews, the final cleaned Emotion dataset consisted of 5,027 entries, and the full cleaning process resulted in the removal of 2,249 entries. Through the dataset cleaning process emotion labels were reduced as follows: "anger" labels from 1,077 to 1,037, "fear" labels from 814 to 787, "happiness" labels from 893 to 789, "hate" labels from 576 to 564, "other" labels from 1,874 to 0, "sadness" labels from 1,158 to 1,042, and "surprise" labels from 884 to 808. The final distribution was 20.6% "anger," 15.7% "fear," 15.7% "happiness," 11.2% "hate," 20.7% "sadness," and 16.1% "surprise."

### 3.3 Data Sampling Strategy

The model evaluation process was made fair through the implementation of a balanced and stratified sampling strategy with guaranteed reproducibility. The Sentiment dataset included three labeling sentiment categories: "negative," "positive," and "neutral." To create a balanced dataset suitable for analysis, 300 texts from each sentiment category were randomly selected in the Google Colab environment. This sampling approach created a final dataset of 900 entries and maintained equal representation for all three sentiment categories.

Similarly, the Emotion dataset consisted of six distinct emotions: "anger," "fear," "happiness," "hate," "sadness," and "surprise." Following the same methodology, 300 texts from each emotion category were randomly selected in the Google Colab. Through this method a balanced dataset containing 1,800 entries was created that ensured equal representation for each emotion category.

Multiple technical controls were implemented throughout the sampling process to ensure perfect reproducibility which allows anyone to generate exactly the same dataset by following the same procedure. These technical controls include: Fixed random seed initialization (value: 42) across all random number generators, deterministic sorting before sampling to create a consistent order, processing emotion/sentiment categories in alphabetical order to maintain identical sequence patterns, and MD5 checksum verification to validate dataset reproducibility. Multiple testing runs confirmed that this methodology produced identical datasets in varied computing environments and execution times.

### 3.4 Model Inference Configuration

#### 3.4.1 Experimental Design and Dataset Preparation

This subchapter describes the setup used to run sentiment analysis and emotion detection tasks using four LLMs which included GPT-4o from OpenAI, Claude 3.7 Sonnet developed by Anthropic, Gemini 2.0 Flash from Google, and DeepSeek-V3 from DeepSeek. All model implementations were carried out using the Google Colab platform. Two balanced datasets were prepared for the evaluation tasks: one dataset includes 900 entries for sentiment analysis and consists of 300 texts per label while the latter has 1,800 entries for emotion detection with 300 texts per label as well.

#### 3.4.2 Prompt Engineering and Standardization

The study required all four LLMs to use the same prompts and input formats for fair evaluations in each task. Consistent task understanding across all four LLMs relied heavily on the prompt design. The early rounds of the study used simple prompt versions which required models to identify the sentiment or emotion directly. The early prompt versions produced hallucinated outputs even though it contained explicit constraints. The research addressed the problem by iteratively testing small balanced datasets containing 120 texts for each task which included 40 texts per sentiment class and 20 texts per emotion category to evaluate and differentiate prompt variants among models.

Automated random sampling selected these samples to provide unbiased coverage. Several prompt iterations were manually evaluated for hallucinations and adherence to the predefined label sets by examining classification report support metrics. The testing process resulted in prompts which optimized phrasing that reduced hallucinations while enforcing strict label usage and delivering consistent results. The design of these prompts established a standard output format in structured data (JSON objects) which assigned each text sample a unique identifier connected to its classified sentiment or emotion. The complete prompts for both sentiment analysis and emotion detection tasks are available in the GitHub repository to ensure full reproducibility.

#### 3.4.3 API Implementation and Batch Processing

A batch size of 20 text entries per API call was adopted to improve the cost-efficiency of API usage and reduce model output errors caused by larger batch sizes. After multiple test runs it was found that larger batch sizes produced hallucinated and non-compliant results especially when the model did not follow the prompt structure or misunderstood label limits.

Through this batching strategy the Sentiment dataset (900 entries) was processed via 45 API calls and the Emotion dataset (1,800 entries) needed 90 API calls per model. The batch processing approach resulted in better error management and minimized occurrences of reaching API rate limits.

### 3.4.4 Evaluation Metrics and Analysis Framework

The evaluation framework used an extensive set of performance metrics to provide multi-dimensional analysis of each model. In addition to basic accuracy measurements, the system produced thorough classification reports which included precision, recall and F1-scores for each emotion and sentiment category. These metrics were calculated using Scikit-learn (a Python machine learning library) to ensure statistical validity.

The confusion matrices created visualizations of misclassification patterns which demonstrated the most common emotion or sentiment pair confusions for each model. Per-category accuracy measurements enabled granular performance analysis, identifying each model's strengths and weaknesses across specific emotions and sentiments.

### 3.4.5 Model Parameter Configuration

Parameters of model inference were controlled so that the evaluations of the LLMs were deterministic and comparable. Temperature was set to 0 for all models, removing any variation in random responses; this also means that model performance can be directly compared on the basis of model capacity rather than random variation in output. A temperature of 0 was especially important for the classification tasks since these results needed to be deterministic.

## 4. Results

This chapter presents the evaluation results of four Large Language Models (LLMs)—Claude 3.7 Sonnet, DeepSeek-V3, Gemini 2.0 Flash, and GPT-4o—on sentiment analysis and emotion detection tasks. The performance of each model is reported through detailed classification reports, accuracy charts, and confusion matrices to identify how each model performs and reveal their strengths and common misclassification patterns across both sentiment analysis and emotion detection tasks.

### 4.1 Sentiment Analysis Results

#### 4.1.1 Claude 3.7 Sonnet Sentiment Analysis

The Claude 3.7 Sonnet model completed processing of the Sentiment dataset with 900 texts in 3 minutes and 55 seconds and incurred a cost of $0.32 which makes it the priciest tested model, and achieved third place in processing speed. The model achieved an overall classification accuracy of 80.42% and a macro-average F1-score of 0.8036. Claude maintained consistent performance across different categories with no extreme weaknesses while achieving second place in sentiment analysis in overall classification accuracy and macro-average F1-score.

|  | precision | recall | f1-score | support |
|---|---|---|---|---|
| negative | 0.85 | 0.78 | 0.81 | 300 |
| neutral | 0.8 | 0.75 | 0.77 | 300 |
| positive | 0.77 | 0.88 | 0.82 | 299 |
| accuracy |  |  | 0.80 | 899 |
| macro avg | 0.81 | 0.8 | 0.8 | 899 |
| weighted avg | 0.81 | 0.8 | 0.8 | 899 |

*Table 2 – Claude Sentiment Analysis Classification Report*

Among all models tested for sentiment analysis Claude produced the highest F1-score for negative sentiment (0.81) which indicates its strong ability to accurately identify genuinely negative content. The model incorrectly assigned "mixed" as the label to text 11 from batch 7 when only negative, neutral, and positive labels were allowed. This mislabeled prediction was excluded from evaluation, resulting in 299 positive texts compared to the expected 300.

The most common misclassification patterns showed that Claude demonstrated difficulty in differentiating neutral expressions from mildly positive ones with 50 instances of neutral-to-positive misclassification (the highest among all models), followed by 39 negative-to-neutral and 28 negative-to-positive confusions.

### 4.1.2 DeepSeek-V3 Sentiment Analysis

DeepSeek-V3 model completed the processing of the 900-text Sentiment dataset in 8 minutes and 33 seconds for $0.01 which gives significant cost benefits although it ranked second behind Gemini, plus it stands as the slowest model among all. Overall classification accuracy achieved was 79.56% and the macro average F1 score was 0.7950, making this model ranked fourth in sentiment analysis among all models.

|  | precision | recall | f1-score | support |
|---:|---:|---:|---:|---:|
| negative | 0.85 | 0.76 | 0.8 | 300 |
| neutral | 0.77 | 0.75 | 0.76 | 300 |
| positive | 0.78 | 0.87 | 0.82 | 300 |
| accuracy |  |  | 0.79 | 900 |
| macro avg | 0.8 | 0.8 | 0.8 | 900 |
| weighted avg | 0.8 | 0.8 | 0.8 | 900 |

*Table 3 - DeepSeek Sentiment Analysis Classification Report*

For sentiment analysis, DeepSeek ranked second (tied with GPT-4o) for negative F1-score, and in general its sentiment analysis performance was only slightly below the other models. The analysis of confusion matrices and pairs shows that the model most frequently misclassified neutral text as positive with 47 occurrences and negative text as neutral with 43 occurrences (both ranked second among all models).

### 4.1.3 Gemini 2.0 Flash Sentiment Analysis

The Gemini 2.0 Flash model completed processing of the Sentiment dataset with 900 texts in 2 minutes and 18 seconds costing less than $0.01 which established its position as the quickest and most economical model evaluated. Overall classification accuracy achieved was 80% and the macro average F1-score was 0.7993, placing the model third in sentiment analysis.

|  | precision | recall | f1-score | support |
|---:|---:|---:|---:|---:|
| negative | 0.85 | 0.74 | 0.79 | 300 |
| neutral | 0.78 | 0.79 | 0.79 | 300 |
| positive | 0.77 | 0.88 | 0.82 | 300 |
| accuracy |  |  | 0.8 | 900 |
| macro avg | 0.8 | 0.8 | 0.8 | 900 |
| weighted avg | 0.8 | 0.8 | 0.8 | 900 |

*Table 4 - Gemini Sentiment Analysis Classification Report*

Gemini achieved the top F1-score for neutral sentiment with a score of 0.79 among all models. However, it had the lowest F1-score for negative sentiment among all other models. The most frequent confusion was between negative and neutral sentiments (42 instances). However, the model showed the highest rate of negative-to-positive misclassification (37 instances) across other models.

### 4.1.4 GPT-4o Sentiment Analysis

The GPT-4o model finished processing the Sentiment dataset which includes 900 texts within a duration of 2 minutes and 56 seconds while costing $0.16. Overall classification accuracy achieved was 80.67% and the macro average F1-score was 0.8070 (both highest among all models). The model showed strong performance across all categories without having any extreme weaknesses in a specific sentiment.

| | precision | recall | f1-score | support |
|---|---|---|---|---|
| negative | 0.85 | 0.74 | 0.8 | 300 |
| neutral | 0.73 | 0.84 | 0.78 | 300 |
| positive | 0.85 | 0.84 | 0.84 | 300 |
| accuracy | | | 0.80 | 900 |
| macro avg | 0.81 | 0.81 | 0.81 | 900 |
| weighted avg | 0.81 | 0.81 | 0.81 | 900 |

*Table 5 – GPT-4o Sentiment Analysis Classification Report*

Across all other models, the GPT-4o model achieved the best F1-score for positive sentiment at 0.84. Analysis of confusion matrices and pairs shows GPT-4o most frequently misinterpreting negative sentences as neutral (58 instances) which leads to a top-ranking error among other models pointing to its inability to identify subtle negative expressions.

### 4.1.5 Statistical Significance Analysis

To validate the statistical significance of the results, bootstrap confidence intervals and McNemar's test were utilized for analyzing the reliability of model performance and significance of pairwise differences. Bootstrapping with 1,000 resamples showed that all models exhibit consistent performance with narrow confidence intervals. Claude reported an accuracy of $0.8047 \pm 0.0132$ (95% CI [0.7775, 0.8309]) and an F1-macro of $0.8038 \pm 0.0132$ (95% CI [0.7766, 0.8292]). DeepSeek achieved an accuracy of $0.7965 \pm 0.0135$ (95% CI [0.7700, 0.8211]) and an F1-macro of $0.7958 \pm 0.0135$ (95% CI [0.7689, 0.8206]). Gemini reached an accuracy of $0.8008 \pm 0.0132$ (95% CI [0.7756, 0.8256]) and an F1-macro of $0.7998 \pm 0.0133$ (95% CI [0.7743, 0.8245]). GPT-4o obtained the highest overall accuracy of $0.8072 \pm 0.0131$ (95% CI [0.7822, 0.8311]) and F1-macro of $0.8073 \pm 0.0130$ (95% CI [0.7822, 0.8310]). The overlapping confidence intervals suggest that while GPT-4o slightly outperforms the other models, the differences are marginal and not decisively significant.

McNemar's test for pairwise model comparison provided evidence in support of the hypothesis: no significant pairwise differences between models were found (all p-values > 0.05). The most significant result was observed between DeepSeek and GPT-4o ($p = 0.3682$), where GPT-4o won 55 times, and DeepSeek 45. Other comparisons, such as Claude and GPT-4o ($p = 0.8376$) and Claude vs Gemini ($p = 0.8099$), also indicate an even distribution of wins and losses across the models. These results confirm that while raw metric values hinted at potential ranking differences, the models' performance on sentiment analysis is not statistically significant.

## 4.2 Emotion Detection Results

### 4.2.1 Claude 3.7 Sonnet Emotion Detection

The Claude 3.7 Sonnet model completed processing of the Emotion dataset with 1,800 texts in 7 minutes and 15 seconds and incurred a cost of $0.71. Similar to the sentiment analysis task, this model was the most expensive one across other models and thus unsuitable for budget-restricted or high-volume usage applications. Overall classification accuracy achieved was 79.27% and the macro average F1-score was 0.7939 (third place for both metrics across other models).

| | precision | recall | f1-score | support |
|---|---|---|---|---|
| anger | 0.65 | 0.8 | 0.72 | 300 |
| fear | 0.91 | 0.89 | 0.9 | 300 |
| happiness | 0.86 | 0.86 | 0.86 | 300 |
| hate | 0.93 | 0.7 | 0.8 | 299 |
| sadness | 0.73 | 0.82 | 0.77 | 300 |
| surprise | 0.75 | 0.68 | 0.71 | 300 |
| accuracy | | | 0.79 | 1799 |
| macro avg | 0.8 | 0.79 | 0.79 | 1799 |
| weighted avg | 0.8 | 0.79 | 0.79 | 1799 |

*Table 6 - Claude Emotion Detection Classification Report*

The model incorrectly assigned "disgust" as the label to one of the texts when only anger, fear, happiness, hate, sadness and surprise labels were allowed. Although disgust can be considered a close semantic relative or synonym of hate, it was not among the predefined valid labels. The evaluation excluded the incorrect prediction which led to a final count of 299 hate-labeled texts rather than the expected 300.

The model achieved top performance in emotion detection for sadness with an F1-score of 0.77 matching GPT-4o and anger with an F1-score of 0.72 matching Gemini. Claude reached peak precision levels for hate speech detection with a score of 0.93 but exhibited the lowest recall rate at 0.70 indicating its conservative approach to marking text as hate speech. In emotion detection, Claude's confusion patterns showed the highest rate of surprise-anger confusions (48 instances), while hate-anger confusions (45 instances) represented the highest rate among all other models.

### 4.2.2 DeepSeek-V3 Emotion Detection

DeepSeek-V3 model completed the processing of 1,800-text Emotion dataset in 16 minutes and 49 seconds for $0.03. As with the sentiment task, the model was very cheap and ranked second in cost-efficiency, but it was the slowest in processing time. Overall classification accuracy achieved was 74.94% and the macro average F1-score was 0.7379 making this model rank fourth again in emotion detection with a more pronounced performance gap in this task compared to sentiment analysis.

|              | precision | recall | f1-score | support |
|--------------|-----------|--------|----------|---------|
| anger        | 0.59      | 0.78   | 0.67     | 300     |
| fear         | 0.88      | 0.95   | 0.91     | 300     |
| happiness    | 0.75      | 0.94   | 0.83     | 300     |
| hate         | 0.8       | 0.74   | 0.77     | 300     |
| sadness      | 0.71      | 0.71   | 0.71     | 300     |
| surprise     | 0.93      | 0.37   | 0.53     | 300     |
| accuracy     |           |        | 0.74     | 1800    |
| macro avg    | 0.77      | 0.75   | 0.74     | 1800    |
| weighted avg | 0.77      | 0.75   | 0.74     | 1800    |

*Table 7 - DeepSeek Emotion Detection Classification Report*

In emotion detection task, it achieved the highest recall for fear (0.95, tied with GPT-4o) and happiness (0.94) among all models, indicating strong capability in identifying these emotions when present. However, DeepSeek's most significant weakness was in surprise detection, where it achieved extremely low recall (0.37), missing 63% of actual surprise expressions. The model demonstrated extensive confusion patterns while detecting emotions especially between surprise and anger where it made 72 mistakes and between surprise and happiness where it made 57 mistakes (both ranked first).

### 4.2.3 Gemini 2.0 Flash Emotion Detection

The Gemini 2.0 Flash model completed processing of the Emotion dataset with 1,800 texts in 4 minutes and 34 seconds costing above $0.01. Similar to its performance in sentiment analysis, it was the fastest and cheapest model among all those models evaluated. Overall classification accuracy achieved was 79.59% and the macro average F1-score was 0.7955. It ranked second overall based on both classification accuracy and macro-average F1-score.

|              | precision | recall | f1-score | support |
|--------------|-----------|--------|----------|---------|
| anger        | 0.64      | 0.83   | 0.72     | 300     |
| fear         | 0.89      | 0.93   | 0.91     | 300     |
| happiness    | 0.84      | 0.88   | 0.86     | 299     |
| hate         | 0.9       | 0.77   | 0.83     | 299     |
| sadness      | 0.75      | 0.75   | 0.75     | 300     |
| surprise     | 0.81      | 0.62   | 0.7      | 300     |
| accuracy     |           |        | 0.79     | 1798    |
| macro avg    | 0.81      | 0.8    | 0.8      | 1798    |
| weighted avg | 0.81      | 0.8    | 0.8      | 1798    |

*Table 8 - Gemini Emotion Detection Classification Report*

The classification of two specific texts (text 17 from batch 4 and text 9 from batch 59) as "shame" and "null" respectively was incorrect because the guidelines stipulated using only six defined labels. The evaluation excluded these two hallucinated predictions—one that should have been labeled as hate but was predicted as shame, and another that should have been labeled as happiness but was predicted as null—leading to a final count of 299 labeled texts instead of the expected 300 for both emotions, due to the use of labels outside the predefined set.

The model reached top F1-score for anger detection at 0.72 which matched Claude's score and secured second place for fear detection at 0.91, happiness detection at 0.86 and hate detection at 0.83 demonstrating strong performance across emotion detection metrics. For emotion detection, Gemini frequently confused surprise with anger (59 instances) and sadness with anger (35 instances), ranked second for both among other models.

### 4.2.4 GPT-4o Emotion Detection

The GPT-4o model finished processing the Emotion dataset which includes 1,800 texts within a duration of 7 minutes and 8 seconds while costing $0.39. Overall classification accuracy achieved was 80.94% and the macro average F1-score was 0.8079. As in the sentiment analysis task, GPT-4o achieved the highest scores and ranked first in both accuracy metrics.

|  | precision | recall | f1-score | support |
|---|---|---|---|---|
| anger | 0.7 | 0.73 | 0.71 | 300 |
| fear | 0.88 | 0.95 | 0.92 | 300 |
| happiness | 0.85 | 0.9 | 0.88 | 300 |
| hate | 0.84 | 0.83 | 0.84 | 300 |
| sadness | 0.79 | 0.75 | 0.77 | 300 |
| surprise | 0.78 | 0.69 | 0.74 | 300 |
| accuracy |  |  | 0.80 | 1800 |
| macro avg | 0.81 | 0.81 | 0.81 | 1800 |
| weighted avg | 0.81 | 0.81 | 0.81 | 1800 |

*Table 9 – GPT-4o Emotion Detection Classification Report*

The emotion detection capabilities of GPT-4o reached peak performance with top F1-scores in fear (0.92), happiness (0.88), hate (0.84), sadness (0.77, tied with Claude), and surprise (0.74) across various emotion categories. While other models demonstrated extreme precision or recall values GPT-4o sustained consistent performance across metrics and achieved the smallest performance disparity between sentiment analysis and emotion detection tasks. The model displayed its highest emotion detection confusion between surprise and anger with 36 instances yet ranked last among models which suggests it struggled to separate these similar emotional states but performed better than others in this regard.

### 4.2.5 Statistical Significance Analysis

To provide further evidence for the reliability of our emotion detection results, bootstrap confidence intervals and McNemar's test were performed. The bootstrap with 1,000 resamples revealed the similar performance of the models with narrow confidence intervals. Claude had an accuracy of $0.7934 \pm 0.0095$ (95% CI [0.7754, 0.8116]) and F1-macro of $0.7944 \pm 0.0094$ (95% CI [0.7762, 0.8125]). The lowest overall performance was of DeepSeek with an accuracy of $0.7499 \pm 0.0102$ (95% CI [0.7294, 0.7689]) and F1-macro of $0.7380 \pm 0.0101$ (95% CI [0.7181, 0.7573]). The performance of Gemini was similar to Claude with an accuracy of $0.7961 \pm 0.0092$ (95% CI [0.7786, 0.8142]) and F1-macro of $0.7955 \pm 0.0091$ (95% CI [0.7776, 0.8132]). GPT-4o had the best scores with an accuracy of $0.8102 \pm 0.0094$ (95% CI [0.7911, 0.8278]) and F1-macro of $0.8083 \pm 0.0094$ (95% CI [0.7896, 0.8265]). The overlapping intervals between Claude, Gemini, and GPT-4o, indicate that although GPT-4o had an advantage, it was rather small. The McNemar's test was run, and results were mixed. All three comparisons with DeepSeek were statistically significant, with Claude outperforming DeepSeek ($p < 0.001$), Gemini outperforming DeepSeek ($p < 0.001$), and GPT-4o also significantly outperforming DeepSeek ($p < 0.001$). However, none of the three comparisons between the higher-scoring models were statistically significant, with Claude vs. Gemini being close to a statistical tie ($p = 0.7439$) and Gemini vs. GPT-4o being close to, but not at, statistical significance ($p = 0.1094$). Claude vs. GPT-4o was also close to statistical significance ($p = 0.0545$), but did not cross the 0.05 threshold.

## 5. Discussion

This chapter conducts a thorough analysis and interpretation of comparative results from evaluating four Large Language Models (LLMs), including Claude 3.7 Sonnet, DeepSeek-V3, Gemini 2.0 Flash, and GPT-4o for sentiment analysis and emotion detection tasks using Persian social media texts. The goal of this section is to interpret and compare the performance of each model against others, discuss its pros and cons, mention some possible justifications for the models' performance, and mention some limitations of this study.

## 5.1 Interpretation of Results

### 5.1.1 Overall Performance Comparison

The evaluated four models show distinct performance variations that are minor but meaningful for both sentiment analysis and emotion detection tasks. The models achieved comparable macro average F1 scores for sentiment analysis with performance metrics extending from DeepSeek-V3's 0.7950 to GPT-4o's 0.8070. The emotion detection task showed model performance macro average F1 scores between 0.7379 for DeepSeek-V3 and 0.8079 for GPT-4o. Sentiment analysis overall macro average F1 score rankings displayed in Figure 1. The highest and lowest macro average F1 scores differed by only 0.0120 which translates to approximately 1.2 percentage points.

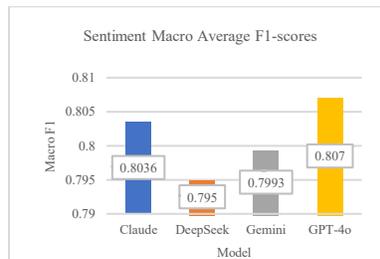

*Figure 1 - Sentiment Analysis Overall Macro Average F1 Score Rankings*

The overall macro average F1 score rankings for the emotion detection task are illustrated in Figure 2. The emotion detection task shows a bigger result gap between the highest and lowest performing models by 0.07 or 7 percentage points. The wider margin proves that differentiating among six emotion categories is basically more challenging than separating three sentiment categories.

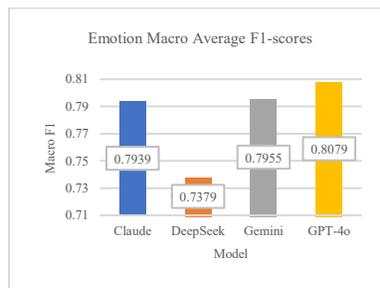

*Figure 2 - Emotion Detection Overall Macro Average F1 Score Rankings*

Performance evaluation should include computational efficiency and cost-effectiveness alongside macro average F1 score metrics. A cost comparison between the four evaluated models for sentiment analysis and emotion detection tasks can be seen in Figure 3. Sentiment analysis costs range between $0.01 and $0.32 and emotion detection costs range between $0.01 and $0.71.

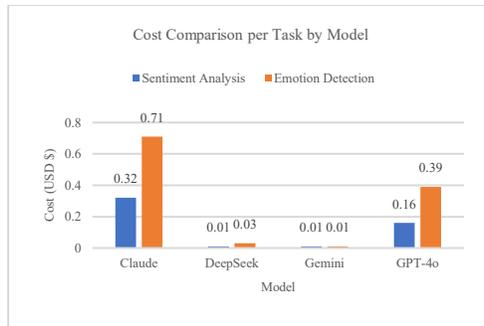

*Figure 3 - Cost Comparison Among LLMs*

Figure 4 shows how processing time analysis uncovers temporal efficiency patterns for different models across various tasks. Sentiment analysis takes from 2 minutes and 18 seconds to 8 minutes and 33 seconds to process while emotion detection processing times range from 4 minutes and 34 seconds to 16 minutes and 49 seconds. The emotion detection task demands longer processing times and higher costs because it analyzes 1,800 texts which is twice as many as the 900 texts in the sentiment analysis dataset.

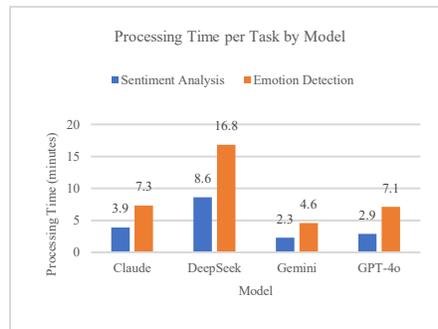

*Figure 4 - Processing Time Analysis Among LLMs*

### 5.1.2 Performance by Task

In sentiment analysis task, all four models showed similar patterns in their performance for all three sentiments. All models demonstrated relatively robust recall for positive sentiment (0.84 for GPT-4o and 0.87-0.88 for others), which indicates that positive expressions in Persian social media texts present distinctive linguistic markers that the models successfully detected. Moreover, the models demonstrated high precision for negative sentiment detection (0.85 across all models) while exhibiting lower recall scores (ranging from 0.74 to 0.78) showing that while models are cautious about labeling text as negative, they fail to detect some genuine negative sentiment expressions. Finally, the neutral sentiment category presented higher difficulties since precision varied between 0.73 (GPT-4o) and 0.80 (Claude) and recall ranged from 0.75 (Claude and DeepSeek) to 0.84 (GPT-4o). Models' performance shows varied ability to interpret expressions lacking clear sentiment indicators. Figure 5 compares F1-score performance across all sentiment categories simultaneously, showing each model's strengths and weaknesses.

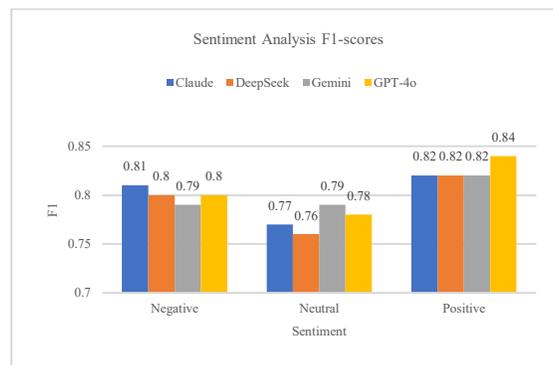

*Figure 5 - F1-Score Performance Across Sentiment Categories Among LLMs*

The emotion detection task showed larger differences between models as well as across various emotion categories. Fear detection demonstrated excellent results across all models with F1-scores ranging between 0.90 for Claude and 0.92 for GPT-4o. The LLMs can easily identify the distinct linguistic patterns used to express fear in Persian. Happiness-related emotional expressions achieved good detection results among models, with F1-scores between 0.83 (DeepSeek) and 0.88 (GPT-4o), indicating that positive emotional expressions were generally well captured.

The performance results for hate detection showed more variation with F1-scores between 0.77 for DeepSeek and 0.84 for GPT-4o. Models showed different precision-recall tradeoffs as precision ranged between 0.80 (DeepSeek) and 0.93 (Claude) and recall varied from 0.70 (Claude) to 0.83 (GPT-4o) which demonstrated their distinct abilities either to avoid wrongly identifying hate content or to fully detect all hate instances. Moreover, among the emotion categories tested, surprise detection proved to be the most difficult task as F1-scores for this classification varied between 0.53 for DeepSeek and 0.74 for GPT-4o. The results of models differed significantly as evidenced by precision scores between 0.75 (Claude) and 0.93 (DeepSeek) and recall scores between 0.37 (DeepSeek) and 0.69 (GPT-4o) which illustrates distinct approaches to Persian surprise detection.

The models achieved moderate results in anger detection as their F1-scores varied between 0.67 for DeepSeek and 0.72 for both Claude and Gemini. Across all models, the precision values for anger were lower than recall values which indicates these models frequently misclassified non-anger text as anger. Finally, models displayed similar performance results in sadness detection as F1-scores varied between 0.71 for DeepSeek and 0.77 for Claude and GPT-4o. The precision scores for models varied moderately between 0.71 for DeepSeek and 0.79 for GPT-4o whereas recall scores for models ranged from 0.71 for DeepSeek to 0.82 for Claude. Both Gemini and DeepSeek showed even results between precision and recall in finding sadness expressions without any leanings toward over-labeling or under-labeling. Figure 6 compares F1-score performance across all emotion categories simultaneously, showing each model's strengths and weaknesses.

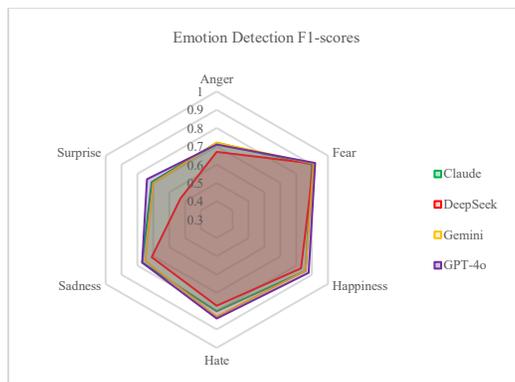

*Figure 6 - F1-Score Performance Across Emotion Categories Among LLMs*

### 5.1.3 Analysis of Common Misclassification Patterns

The analysis of confusion pairs among different models demonstrates several persistent patterns that highlight the inherent challenges associated with sentiment analysis in Persian texts. Across all models the most common misclassification was between negative and neutral sentiments which appeared 182 times. The extensive overlap demonstrates that particular challenges exist when trying to differentiate mild negativity from neutral expressions in Persian social media texts. The second most frequent confusion, observed 163 times, was between neutral and positive sentiments. The pattern shows that models frequently fail to distinguish between true positive expressions and neutral statements which include positively connoted words but do not express strong emotions. A third prominent confusion, with 112 instances, occurred between negative and positive sentiments. The challenge emerges from expressions that use positive language to disguise negative content through mixed or indirect emotional signals. Figure 7 shows how these confusion patterns appear among various LLMs.

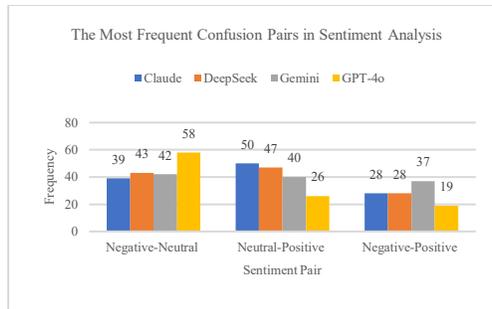

*Figure 7 - Sentiment Analysis Confusion Patterns Among LLMs*

For emotion detection, several recurring misclassification patterns emerged across models, each revealing important challenges in distinguishing emotional expressions in Persian social media texts. Across all models there were 215 instances where surprise was confused with anger as the most frequent misclassification. Computational models face difficulties in capturing subtle contextual signals which result in surprise expressions being frequently misinterpreted as anger. The semantic closeness between hate and anger caused 152 instances of confusion because it demonstrates how challenging it is to distinguish varying negative emotional expressions and intensities. The frequent confusion between sadness and anger which appeared 117 times suggests that the Persian language uses overlapping linguistic elements like vocabulary and syntax to describe these emotions. In addition, a total of 117 times surprise was mistaken for happiness likely due to the fact that surprise possesses an ambivalent nature that allows it to express positive emotions depending on the situation. Finally, the 84 cases where surprise was confused with sadness, further show how surprise can contextually express negative emotions and reveal its ambiguous nature. Figure 8 provides an illustration of how confusion pairs are distributed among LLMs.

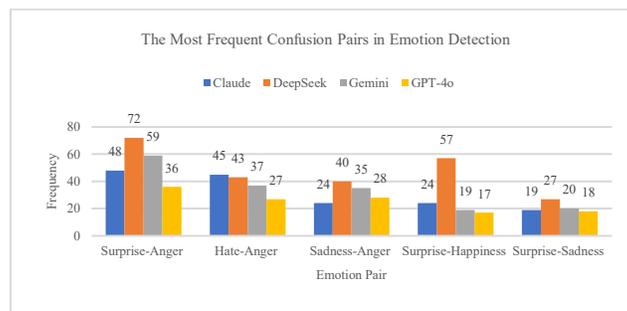

*Figure 8 - Emotion Detection Confusion Pairs Distribution Among LLMs*

### 5.2 Implications for Persian NLP Applications

LLMs show promising results in Persian sentiment analysis and emotion detection according to this comparative study which establishes essential guidelines for creating Persian NLP applications. Modern LLMs achieve strong performance when analyzing Persian texts with results showing macro average F1-scores between 73.79% and 80.79% across various models and tasks which proves their practical utility for Persian NLP applications.

Model performance profiles for both sentiment and emotion tasks have distinct patterns, suggesting the need for task-aware model selection in implementation strategies. The efficiency-accuracy tradeoff is another important consideration in which time-sensitive or cost-constrained applications benefit from Gemini's time-to-cost ratio while applications with high accuracy demands should choose GPT-4o even at a greater computational cost. Variances in speed and cost among models provide clear guidance for resource allocation across different deployment contexts.

The performance differences across emotion categories in the models including consistent difficulty with surprise detection and better recognition of fear and happiness will affect applications working with sensitive content or needing balanced emotion detection. The uneven performance among emotion categories can produce biased results when used in applications such as mental health monitoring or customer feedback analysis which require thorough attention to fairness and equity throughout implementation.

The models perform well in detecting negative sentiment and anger which enables customer support systems to flag unhappy customers in need of immediate assistance. The batch processing approach used in this paper was cost-effective but showed potential limitations for practical applications requiring processing of individual texts for the best accuracy.

Results are driven by social media data which suggests conclusions are most relevant to the field of processing informal text, and may not effectively apply to formal writing or academic content analysis. Domain specificity requires organizations to thoroughly assess model performance using their unique text types prior to implementation. The subjective interpretation inherent in emotion and sentiment annotation drives the requirement for strong evaluation frameworks and the potential inclusion of human-in-the-loop systems in essential applications.

Modern LLM technology enables successful commercial Persian language processing applications since organizations can choose models that meet their particular needs regarding accuracy, cost, and speed.

## 6. Conclusion

The comparison of the four LLMs on the Persian datasets for the sentiment analysis and emotion detection tasks revealed some interesting findings and enhance knowledge about LLM effectiveness on Persian social media texts. The GPT-4o model, with the highest accuracy (80.67% and 80.94% for sentiment analysis and emotion detection, respectively) among the other models, also exhibited a fairly balanced performance between precision and recall for all categories. Gemini 2.0 Flash was found to be the cheapest and the fastest model, with competitive accuracy (80% for sentiment, 79.59% for emotion). Claude 3.7 Sonnet had the best precision scores and especially high precision in hate detection at 93%. DeepSeek-V3, despite its lower cost, underperforms in comparison to others, especially in emotion detection, with a lower accuracy rate of 74.94% and longer response times.

The models had more difficulties with emotion detection than with sentiment analysis in all cases, with a 7 percentage point performance difference as compared to 1.2 percentage points. Although all models showed sufficient sentiment analysis capabilities on Persian social media texts, the model performance in emotion detection showed a larger spread when classifying into the six emotion classes. All models performed well in detecting fear (F1: 0.90-0.92) while they were less capable in surprise detection, where F1-scores ranged from 0.53 (DeepSeek) to 0.74 (GPT-4o).

Clear tradeoffs were seen between performance and resource consumption. Gemini demonstrated the most efficiency with processing times under 7 minutes and costs below $0.02, while maintaining competitive accuracy. GPT-4o had the highest accuracy despite higher expenses ($0.16-$0.39), thus suitable for high-performance applications. Claude exhibited solid accuracy at a premium price ($0.32 - $0.71) and can be considered for accuracy prioritized applications. DeepSeek had the lowest costs ($0.01-$0.03), but with severe time penalties (8-17 minutes) and lower emotion detection performance.

The analysis of the systematic misclassification patterns also exposed some shortcomings that go beyond individual models when it comes to sentiment and emotion analysis in Persian. The most common confusions in sentiment analysis across all models were between negative and neutral sentiments (182 cases across all models), which could point towards a potential difficulty in distinguishing between slightly negative and neutral content in Persian social media texts. In terms of emotion detection, the predominant type of confusion was between surprise-anger (215 cases) followed by hate-anger (152 cases) and sadness-anger (117 cases). These patterns indicate potential semantic overlap in Persian linguistic expressions for these emotions.

Some constraints affected the findings and the generalizability of the study. The use of batch processing for cost reasons may have affected the quality of responses compared to individual API calls. The datasets only included social media texts, so the generalizability of the results to other domains of the Persian language may be limited. Sentiment and emotion labeling is a subjective task, and the evaluation metric only accounted for binary correctness without considering partial correctness or prediction confidence. Models were evaluated in a zero-shot setting without access to any context, and the input was only a standalone piece of text without any additional conversational, social, or temporal information that could aid in the analysis.

Organizations planning to deploy LLMs for Persian text analysis can make informed decisions through these findings by considering their operational requirements and financial constraints. Although current LLMs show promising performance for use in real applications of Persian sentiment analysis and emotion detection, deployment would require evaluation on the intended task, and it could be improved by using hybrid approaches that combine LLMs with fine-tuned models in the domain to handle the problem of low emotion categories and confusion.